\pdfoutput=1
\documentclass[11pt]{article}

\usepackage[]{acl}

\usepackage{xcolor}

\usepackage{times}
\usepackage{latexsym}
\usepackage{amsmath}
\usepackage[T1]{fontenc}
\usepackage{pgfplots}

\pgfplotsset{width=10cm,compat=1.9}

\usepackage[utf8]{inputenc}
\usepackage{graphicx}
\usepackage{times}
\usepackage{latexsym}
\usepackage{courier}

\usepackage{amssymb}
\usepackage{stmaryrd}
\usepackage{tikz}
\usepackage{tikz-qtree}
\usepackage{booktabs}
\usepackage{xcolor}
\usepackage{makecell}
\usepackage{amsfonts}
\usepackage{booktabs}
\usepackage{siunitx}
\usepackage{tabularx}
\usepackage{multirow}
\newcommand{\comment}[1]{}
\usepackage{listings}
\usepackage{xparse}
\usepackage{float}

\NewDocumentCommand{\codeword}{v}{%
\texttt{{#1}}%
}

\newcommand{\var}{\sc}

\usepackage{graphicx}
\graphicspath{ {./images/} }

\usepackage[T1]{fontenc}
\usepackage[utf8]{inputenc}

\usepackage{microtype}
\usepackage{ulem}
\newcolumntype{Y}{>{\centering\arraybackslash}X}

\usepackage{microtype}

\newcommand{\astc}[1]{{\bf\color{red!50} #1}}
\newcommand{\astf}[1]{{\em\color{blue!30!green!70} #1}}
\newcommand{\asts}[1]{{\color{blue!70} #1}}

\title{The impact of lexical and grammatical processing on generating code from natural language}

\author{Nathanaël Beau$^{1,2}$ \and Benoît Crabbé$^1$ \\
       $^1$ Université de Paris, LLF, CNRS, 75013 Paris, France \\ $^2$ onepoint, 29 rue des Sablons, F-75116 Paris, France \\
       \texttt{n.beau@groupeonepoint.com} \\
       \texttt{benoit.crabbe@u-paris.fr} }

\begin{document}
\maketitle
\begin{abstract}

Considering the seq2seq architecture of \citet{tranx-2018} for natural language to code translation, we identify four key components of importance: grammatical constraints, lexical preprocessing, input representations, and copy mechanisms. To study the impact of these components, we use a state-of-the-art architecture that relies on BERT encoder and a grammar-based decoder for which a formalization is provided. The paper highlights the importance of the lexical substitution component in the current natural language to code systems.

% \sout{We present the contribution of the BERT encoder to the problem of semantic parsing. We aim at translating natural language into a formal meaning representation in Python. We characterise the real syntactic contribution in this translation task on two data sets. We show the limits of the conclusions that can be drawn by analysing the semantic features of the data sets, describing the impact of preprocessing on the BLEU score and highlighting the biases contained in the Python generation data sets.}
\end{abstract}

\section{Introduction}

Translating natural language program descriptions to actual code is meant to help programmers to ease writing reliable code efficiently by means of a set of advanced code completion mechanisms.

% \sout{Communicating with a machine is one of humanity's dreams. As early as 1950, Turing published his famous article \cite{Turing} testing the ability of a machine to imitate a human conversation. The problem of translating natural language into a formal language intelligible to the machine is a step in this direction. }

There are mainly two classes of methods for obtaining code corresponding to a query expressed in natural language. The first one is code retrieval, which consists of searching and retrieving an appropriate code snippet from a code database. 
%A large amount of data us available online on GitHub or Stackoverflow \cite{CodeSearchNet}, thus it is likely that the desired code is readily available in the database, the difficulty here being to select an appropriate element in a vast set of possibilities \cite{MicrosoftRetrieval}.
The second one is code generation, where the goal is to generate code fragments from a natural language description, generating potentially previously unseen code. In this work, we are interested in Python code generation. Code generation features a mismatch between an ambiguous and noisy natural language input and the structured nature of the generated code. Although Python's vocabulary has a finite number of keywords, the set of values that can be assigned to a variable is infinite and constitutes one of the issues in predicting code corresponding to natural language.

Like many other NLP tasks, current architectures for natural language to code generally take advantage  of pre-trained language models such as BERT \cite{BERT} or GPT \cite{GPT} based on the transformer architecture \cite{Transformer}.  In particular, these architectures are used for code generation where parallel data is limited due to the human expertise required for alignment. The best results on code generation are reached by pretraining seq2seq models on external sources, then by fine-tuning those models on smaller data sets. For instance, \citet{orlanski-21} fine-tune BART \cite{BART} on data pairs of natural language and code and by taking advantage of external informations. Similarly, \citet{monolingualdata} used BERT and a transformer decoder in a semi-supervised way by taking advantage of a large amount of additional monolingual data. Another popular method is to train large language models on code \cite{LLM1, APPS}. Notably, GPT-3 has been finetuned on a large quantity of data from Github to obtain a powerful language model named Codex \cite{Codex} that powers Github Copilot, a tool to help developers. 

Overall the above mentioned solutions aim to take advantage of large amounts of training data available nowadays, but few of them care about generating code that is guaranteed to be syntactically correct nor well typed. Let us mention some exceptions from semantic parsing like \citet{LSTMtree, ASN,tranx-1} that rely on grammatical constraints to ensure that the generated code can be executable. 

In this work, we study variations around the {\fontfamily{cmss}\selectfont TranX} seq2seq architecture \citep{tranx-2018} for translating natural language to code. 
Rather than generating directly code tokens from natural language, the architecture generates an Abstract Syntax Tree (AST) constrained by the programming language grammar. %{\fontfamily{cmss}\selectfont TranX} extends the AST at each step during decoding, which slows down the decoding process. 

The paper reports state of the art results on the task and specifically introduces:
\begin{itemize}
    \item A formalization of the grammar constrained code generator relying on the \citet{earley:1970} parser transition system.
    \item A study of the impact of key components of the architecture on the performance of the system: we study the impact of the grammatical component itself, the impact of the language model chosen, the impact of variable naming and typing and the impact of the input/output copy mechanisms.
\end{itemize}
It is structured as follows. Section {\ref{TransitionSystem}} formalizes the symbolic transition system used for generating the grammatically correct code, Section \ref{Code Generation} describes a family of variants around the {\fontfamily{cmss}\selectfont TranX} architecture that will be used to study the impact of these variations in the experimental part of the paper (Section \ref{Experiments}).

\section{A transition system for code generation} \label{TransitionSystem}

\begin{figure*}[htbp]
\begin{center}
\hspace{-0.5cm}\scalebox{0.77}{\tt \begin{tabular}{lcl}
 \asts{expr} &=& \astc{BinOp} \asts{expr} \astf{left},
 \asts{operator} \astf{op}, \asts{expr} \astf{right}\\
\asts{operator} &=& \astc{Add}\\
\asts{expr} &=& \astc{Constant} \asts{constant} \astf{value}\\
\asts{expr} &=&\astc{List} \asts{expr*} \astf{elts}
\end{tabular}}
\end{center}
\caption{\label{fig:asdl}Example of ASDL rules for the Python language. Each rule is built from a set of grammatical symbols (in blue), is uniquely identified by a constructor name (in red) and provides names to its right hand side symbols, its fields (in green). Grammatical symbols are split in nonterminals (like {\tt expr}) and terminals or primitives (like {\tt constant}). Grammatical symbols can also be annotated with qualifiers ({\tt *}) that allow for zero or more iterations of the symbol.}
\end{figure*}

% In this section we describe a method that ensures on the one hand that the generated code is syntactically valid and on the other hand that the whole translation process reduces to a {\em seq2seq} problem in order to take advantage of current deep learning architectures for {\sc nlp}.

Among the models tested in the paper, some are generating syntactically constrained code. In the context of our study, we propose a transition model that meets two objectives: the code generated is grammatically valid in terms of syntax and the whole translation process still reduces to a seq2seq transduction mechanism that allows us to leverage standard machine learning methods.

% We describe here a transition system which ensures on the one hand that the generated code is syntactically valid and on the other hand that the whole translation process still reduces to a seq2seq problem in order to take advantage of current deep learning architectures for {\sc nlp}.

To this end we introduce a transition system for code generation that generates an AST as a sequence of actions. 
The derivations can then be translated into ASTs and in actual Python code by means of deterministic functions. The set of valid ASTs is a set of trees that are generated by an ASDL grammar \cite{Wang:97}. An ASDL grammar is essentially a context free grammar abstracting away from low level syntactic details of the programming language and aims to ease the semantic interpretation of the parse trees. To this end ASDL grammar rules come  with additional decorators called constructors and field names (Figure \ref{fig:asdl}). 

Our transition system generates derivations, or sequences of actions, that can be translated to a syntactically correct Python code. We adapt to code generation the transition system of the Earley parser \cite{earley:1970} as formalized in Figure \ref{fig:earley}. The generator state is a stack of dotted rules. A dotted rule is a rule of the form $A\rightarrow \alpha \bullet X \beta$ where $\alpha$ is a sequence of grammar symbols whose subtrees are already generated and $X\beta$ is a sequence of grammar symbols for which the subtrees are yet to be generated. The $\bullet X$ symbol is the dotted symbol or the next symbol for which the system has to generate the subtree. The Python ASDL grammar includes rules with star ($*$) qualifiers allowing zero or more occurrences of the starred symbol. The transition system uses an additional set of starred actions and a {\sc close} action to stop these iterations (Figure \ref{fig:earley}).  

Each {\sc predict}(C) action starts the generation of a new subtree from its parent. The {\sc generate} action adds a new leaf to a tree. The {\sc complete} action finishes the generation of a subtree and continues the generation process with its parent. The set of {\sc predict} actions is parametrized by the ASDL rule constructor ($C$), thus there are as many predict actions as there are constructors in the ASDL grammar. Constructors are required in order to generate the actual ASTs from the derivations.

{\sc generate}(V) actions are actions responsible for generating the terminal or primitive symbols. The Python ASDL grammar generates ASTs with primitive leaf types ({\tt identifier, int, string, constant}) that have to be filled with actual values for the AST to be useful. 
To generate actual primitive values the set of generate actions is also parametrized by the actual values $V$ for the primitive types. The set of such values is infinite and consequently the set of generate actions is also infinite. 

Non-Determinism comes from the use of {\sc predict}(C), {\sc generate}(V) and {\sc close} rules. By contrast the application of the {\sc complete} action is entirely deterministic: once the generator has a completed dotted rule on the top of its stack, it has no other choice than applying the complete rule.

\begin{figure*}
\begin{center}
\begin{minipage}{.57\textwidth}
\scalebox{0.6}
{\begin{tabular}{ll}\toprule
{\bf Generator State (stack)}&{\bf Action}\\\midrule
$\langle \mathtt{expr} \rightarrow \bullet\mathtt{expr}^* \rangle$&{\sc start}(List)\\
$\langle \mathtt{expr} \rightarrow \bullet\mathtt{expr}^* |\mathtt{expr}\rightarrow \bullet \mathtt{expr}\, \mathtt{operator}\,\mathtt{expr} \rangle$&{\sc predict}$^*$(BinOp)\\
$\langle \mathtt{expr} \rightarrow \bullet\mathtt{expr}^* |\mathtt{expr}\rightarrow \bullet \mathtt{expr}\, \mathtt{operator}\,\mathtt{expr}| \mathtt{expr} \rightarrow\bullet \mathtt{constant}\rangle$&{\sc predict}(Constant)\\
$\langle \mathtt{expr} \rightarrow \bullet\mathtt{expr}^* |\mathtt{expr}\rightarrow \bullet \mathtt{expr}\, \mathtt{operator}\,\mathtt{expr}| \mathtt{expr} \rightarrow \mathtt{constant}\bullet\rangle$&{\sc Generate}(7)\\
$\langle \mathtt{expr} \rightarrow \bullet\mathtt{expr}^* |\mathtt{expr}\rightarrow \mathtt{expr}\, \bullet\mathtt{operator}\,\mathtt{expr}\rangle $&{\sc complete}\\
$\langle \mathtt{expr} \rightarrow \bullet\mathtt{expr}^* |\mathtt{expr}\rightarrow \mathtt{expr}\, \bullet\mathtt{operator}\,\mathtt{expr} | \mathtt{expr}\rightarrow\bullet\rangle $&{\sc predict}(Add)\\
$\langle \mathtt{expr} \rightarrow \bullet\mathtt{expr}^* |\mathtt{expr}\rightarrow \mathtt{expr}\, \mathtt{operator}\,\bullet\mathtt{expr} \rangle $&{\sc complete}\\
$\langle \mathtt{expr} \rightarrow \bullet\mathtt{expr}^* |\mathtt{expr}\rightarrow \mathtt{expr}\, \mathtt{operator}\,\bullet\mathtt{expr}|\mathtt{expr}\rightarrow \bullet\mathtt{constant}\rangle $&{\sc predict}(Constant)\\
$\langle \mathtt{expr} \rightarrow \bullet\mathtt{expr}^* |\mathtt{expr}\rightarrow \mathtt{expr}\, \mathtt{operator}\,\bullet\mathtt{expr}|\mathtt{expr}\rightarrow \mathtt{constant}\bullet\rangle $&{\sc generate}(5)\\
$\langle \mathtt{expr} \rightarrow \bullet\mathtt{expr}^* |\mathtt{expr}\rightarrow \mathtt{expr}\, \mathtt{operator}\,\mathtt{expr}\bullet\rangle $&{\sc complete}\\
$\langle \mathtt{expr} \rightarrow \mathtt{expr}\bullet\mathtt{expr}^* \rangle $&{\sc complete}$^*$\\
%$\langle \mathtt{expr} \rightarrow \bullet\mathtt{expr}^* |\mathtt{expr}\rightarrow \mathtt{expr}\, \mathtt{operator}\,\mathtt{expr}\bullet\rangle $&{\sc complete}\\
$\langle \mathtt{expr} \rightarrow \mathtt{expr}\bullet\mathtt{expr}^* | \mathtt{expr}\rightarrow\bullet\mathtt{constant}\rangle $&{\sc predict}$^*$(Constant)\\
$\langle \mathtt{expr} \rightarrow \mathtt{expr}\bullet\mathtt{expr}^* | \mathtt{expr}\rightarrow\mathtt{constant}\bullet\rangle $&{\sc generate}(4)\\
$\langle \mathtt{expr} \rightarrow \mathtt{expr}\,\mathtt{expr}\bullet\mathtt{expr}^* \rangle $&{\sc complete}$^*$\\
$\langle \mathtt{expr} \rightarrow \mathtt{expr}\,\mathtt{expr}\bullet\rangle $&{\sc close}$^*$\\\bottomrule
\end{tabular}
}
\end{minipage}
\begin{minipage}{.4\textwidth}
\tikzset{every tree node/.style={align=center,anchor=north}}
\tikzset{level distance=40pt}
\scalebox{0.65}{
\begin{tikzpicture}
\Tree [.{\asts{expr}\\\astc{(List)}} [.\node(left){\asts{expr}\astf{:elts}\\\astc{(BinOp)}}; [.{\asts{expr}\astf{:left}\\\astc{(Constant)}} {\asts{constant}\astf{:value}\\{\bf 7}} ] {\asts{operator}\astf{:op}\\\astc{(Add)}} [.{\asts{expr}\astf{:right}\\\astc{(Constant)}} {\asts{constant}\astf{:value}\\{\bf 5}} ] ]   [.\node(right){\asts{expr}\astf{:elts}\\\astc{(Constant)}}; {\asts{constant}\astf{:value}\\{\bf 4}} ] ]
\fill [rounded corners=0.3cm,fill=yellow, fill opacity=0.2] (left.south west) rectangle (right.north east);
\end{tikzpicture}
}
\end{minipage}

\end{center}
\caption{\label{fig:example}Example derivation for the generation of the Python list expression {\tt [7+5,4]}. The derivation starts with $\mathtt{expr}$ as axiom symbol and applies transitions until the goal is reached. The list of actions performed is called the generator {\bf derivation}. Given a generated derivation we can design a straightforward deterministic procedure to translate it into an {\sc ast}. The actual Python code is generated from the {\sc ast} by the {\tt astor} library.}
\end{figure*}
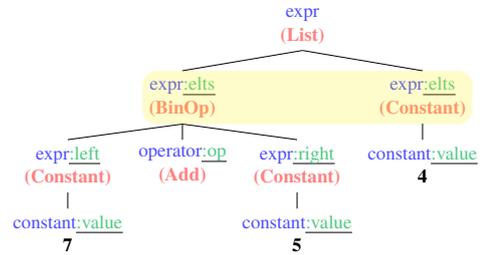

The sequential generation process is illustrated in Figure \ref{fig:example}. Given a start state, at each time step, the generator has to decide which action to perform according to the current state of the stack and updates the stack accordingly. Once the generator reaches the goal state, we collect the list of actions performed (the derivation) in order to build the AST that we finally translate into actual Python code\footnote{We use the \texttt{astor} library to this end.}.

\begin{figure*}
\begin{center}
\scalebox{0.95}{
\begin{tabular}{llcll}\toprule
{\bf Action}&{\bf Transition}&&&{\bf Condition}\\\midrule
{\sc start}(C)&$\langle A \rightarrow \bullet \alpha \rangle$\\
{\sc goal }&$\langle A\rightarrow  \alpha\bullet \rangle$\\\midrule
{\sc predict}(C)&$\langle\mathbf{S}| A\rightarrow \alpha\bullet B\beta \rangle$& $\Rightarrow$& $\langle \mathbf{S}| A\rightarrow \alpha\bullet B\beta  | B \rightarrow \bullet \gamma \rangle$&$(B\rightarrow \gamma \in \text{rules})$\\
{\sc generate}(V)&$\langle\mathbf{S}| A\rightarrow \alpha\bullet t \beta \rangle$&$ \Rightarrow$&$ \langle \mathbf{S}|A\rightarrow \alpha t\bullet\beta  \rangle$&$(t\in \text{primitives})$\\
{\sc complete}&$\langle\mathbf{S}| A\rightarrow \alpha\bullet B\beta |B\rightarrow  \gamma  \bullet\rangle$& $\Rightarrow $&$\langle\mathbf{S}| A\rightarrow \alpha B \bullet \beta\rangle$\\\midrule
{\sc predict}$^*$(C)&$\langle\mathbf{S}| A\rightarrow \alpha\bullet B^*\beta \rangle$& $\Rightarrow$& $\langle \mathbf{S}| A\rightarrow \alpha \bullet B^*\beta  | B \rightarrow \bullet \gamma\rangle$&$(B\rightarrow \gamma \in \text{rules})$\\
{\sc generate}$^*$(V)&$\langle\mathbf{S}| A\rightarrow \alpha\bullet t^* \beta \rangle$ &$\Rightarrow$&$\langle \mathbf{S}|A\rightarrow \alpha t^\bullet t^*\beta  \rangle$&$(t\in \text{primitives})$\\
{\sc complete}$^*$&$\langle\mathbf{S}| A\rightarrow \alpha\bullet B^*\beta |B\rightarrow  \gamma  \bullet\rangle $&$\Rightarrow$&$ \langle\mathbf{S}| A\rightarrow \alpha B\bullet B^* \beta\rangle$\\
{\sc close}$^*$ &$\langle\mathbf{S}|A\rightarrow \alpha \bullet X^*\beta \rangle$&$\Rightarrow$& $\langle\mathbf{S}|A\rightarrow \alpha \bullet\beta \rangle$\\\bottomrule
\end{tabular}}
\end{center}
\caption{\label{fig:earley} An Earley inspired transition system for generating Abstract Syntactic Trees. The state of the generator is a  stack of dotted rules whose bottom is $\mathbf{S}$. As in the the Earley parser, the {\sc predict} rule starts the generation of a new subtree by pushing a new dotted rule on the stack, the {\sc generate} rule adds a leaf to the tree by swapping the top of the stack and the {\sc complete} rule attaches a generated subtree into its parent by popping the top two elements of the stack and pushing an updated dotted rule.
To handle {\tt *} qualifiers we add the starred inference rules where {\sc complete}$^*$ and {\sc generate}$^*$ implement an iteration that stops with the {\sc close}$^*$ rule. 
}
\end{figure*}

\comment{
\begin{figure*}
\begin{center}
\begin{minipage}{.57\textwidth}
\scalebox{0.6}
{\begin{tabular}{ll}\toprule
{\bf Generator State (stack)}&{\bf Action}\\\midrule
$\langle \mathtt{expr} \rightarrow \bullet\mathtt{expr} \rangle$&{\sc start}(Expr)\\
$\langle \mathtt{expr} \rightarrow \bullet\mathtt{expr} |\mathtt{expr}\rightarrow \bullet \mathtt{expr}\, \mathtt{expr}^*\,\mathtt{keyword}^* \rangle$&{\sc predict}(Call)\\
$\langle \mathtt{expr} \rightarrow \bullet\mathtt{expr} |\mathtt{expr}\rightarrow \bullet \mathtt{expr}\, \mathtt{expr}^*\,\mathtt{keyword}^*| \mathtt{expr} \rightarrow\bullet \mathtt{identifier}\ \mathtt{expr\_context}\rangle$&{\sc predict}(Name)\\
$\langle \mathtt{expr} \rightarrow \bullet\mathtt{expr} |\mathtt{expr}\rightarrow \bullet \mathtt{expr}\, \mathtt{expr}^*\,\mathtt{keyword}^*| \mathtt{expr} \rightarrow\bullet \mathtt{identifier}\ \mathtt{expr\_context}\rangle$&{\sc generate}(foo)\\
$\langle \mathtt{expr} \rightarrow \bullet\mathtt{expr}^* |\mathtt{expr}\rightarrow \bullet \mathtt{expr}\, \mathtt{operator}\,\mathtt{expr}| \mathtt{expr} \rightarrow \mathtt{constant}\bullet\rangle$&{\sc Generate}(7)\\
$\langle \mathtt{expr} \rightarrow \bullet\mathtt{expr}^* |\mathtt{expr}\rightarrow \mathtt{expr}\, \bullet\mathtt{operator}\,\mathtt{expr}\rangle $&{\sc complete}\\
$\langle \mathtt{expr} \rightarrow \bullet\mathtt{expr}^* |\mathtt{expr}\rightarrow \mathtt{expr}\, \bullet\mathtt{operator}\,\mathtt{expr} | \mathtt{expr}\rightarrow\bullet\rangle $&{\sc predict}(Add)\\
$\langle \mathtt{expr} \rightarrow \bullet\mathtt{expr}^* |\mathtt{expr}\rightarrow \mathtt{expr}\, \mathtt{operator}\,\bullet\mathtt{expr} \rangle $&{\sc complete}\\
$\langle \mathtt{expr} \rightarrow \bullet\mathtt{expr}^* |\mathtt{expr}\rightarrow \mathtt{expr}\, \mathtt{operator}\,\bullet\mathtt{expr}|\mathtt{expr}\rightarrow \bullet\mathtt{constant}\rangle $&{\sc predict}(Constant)\\
$\langle \mathtt{expr} \rightarrow \bullet\mathtt{expr}^* |\mathtt{expr}\rightarrow \mathtt{expr}\, \mathtt{operator}\,\bullet\mathtt{expr}|\mathtt{expr}\rightarrow \mathtt{constant}\bullet\rangle $&{\sc generate}(5)\\
$\langle \mathtt{expr} \rightarrow \bullet\mathtt{expr}^* |\mathtt{expr}\rightarrow \mathtt{expr}\, \mathtt{operator}\,\mathtt{expr}\bullet\rangle $&{\sc complete}\\
$\langle \mathtt{expr} \rightarrow \mathtt{expr}\bullet\mathtt{expr}^* \rangle $&{\sc complete}$^*$\\
%$\langle \mathtt{expr} \rightarrow \bullet\mathtt{expr}^* |\mathtt{expr}\rightarrow \mathtt{expr}\, \mathtt{operator}\,\mathtt{expr}\bullet\rangle $&{\sc complete}\\
$\langle \mathtt{expr} \rightarrow \mathtt{expr}\bullet\mathtt{expr}^* | \mathtt{expr}\rightarrow\bullet\mathtt{constant}\rangle $&{\sc predict}$^*$(Constant)\\
$\langle \mathtt{expr} \rightarrow \mathtt{expr}\bullet\mathtt{expr}^* | \mathtt{expr}\rightarrow\mathtt{constant}\bullet\rangle $&{\sc generate}(4)\\
$\langle \mathtt{expr} \rightarrow \mathtt{expr}\,\mathtt{expr}\bullet\mathtt{expr}^* \rangle $&{\sc complete}$^*$\\
$\langle \mathtt{expr} \rightarrow \mathtt{expr}\,\mathtt{expr}\bullet\rangle $&{\sc close}$^*$\\\bottomrule
\end{tabular}
}
\end{minipage}
\begin{minipage}{.4\textwidth}
\tikzset{every tree node/.style={align=center,anchor=north}}
\tikzset{level distance=40pt}
\scalebox{0.45}{
\begin{tikzpicture}
\Tree [.{\asts{expr}\\\astc{(Expr)}} [.\node(left){\asts{expr}\astf{:value}\\\astc{(Call)}}; [.{\asts{expr}\astf{:func}\\\astc{(Attribute)}}
[.{\asts{expr}\astf{:value}\\\astc{(Name)}} 
{\asts{identifier}\astf{:id}\\{\bf foo}}
{\asts{expr\_context}\astf{}\\\astc{(Load)}} ]
{\asts{identifier}\astf{:attribute}\\{\bf append}} {\asts{expr\_context}\astf{}\\\astc{(Load)}} ] [.{\asts{expr}\astf{:args}\\\astc{(List)}}
[.{\asts{expr}\astf{:value}\\\astc{(Num)}} 
{\asts{object}\astf{:n}\\{\bf 8}} ]
[.{\asts{expr}\astf{:value}\\\astc{(Num)}} 
{\asts{object}\astf{:n}\\{\bf 7}} ]
{\asts{expr\_context}\astf{}\\\astc{(Load)}} ] {\asts{expr\_context}\astf{}\\\astc{(Load)}} ] ]
\fill [rounded corners=0.3cm,fill=yellow, fill opacity=0.2] (left.south west) rectangle (right.north east);
\end{tikzpicture}
}
\end{minipage}

\end{center}
\caption{\label{fig:exampleNath}Example derivation NATH for the generation of the Python list expression {\tt foo.append([8,7])}. The derivation starts with $\mathtt{expr}$ as axiom symbol and applies transitions until the goal is reached. The list of actions performed is called the generator {\bf derivation}. Given a generated derivation we can design a straightforward deterministic procedure to translate it into an {\sc ast}. The actual Python code is generated from the {\sc ast} by the {\tt astor} library}
\end{figure*}
}

\section{Factors influencing code prediction}
\label{Code Generation}

All architectures analyzed in this study are variations around a seq2seq architecture. We describe the several variants of this architecture used in this paper both on the encoder and decoder side.
We identify key factors that have an impact on the natural-language-to-code translation architecture and we formalize a family of models that allow to test variations of these factors.

\begin{figure*}
\begin{center}
\scalebox{0.9}{
\includegraphics[width=\textwidth]{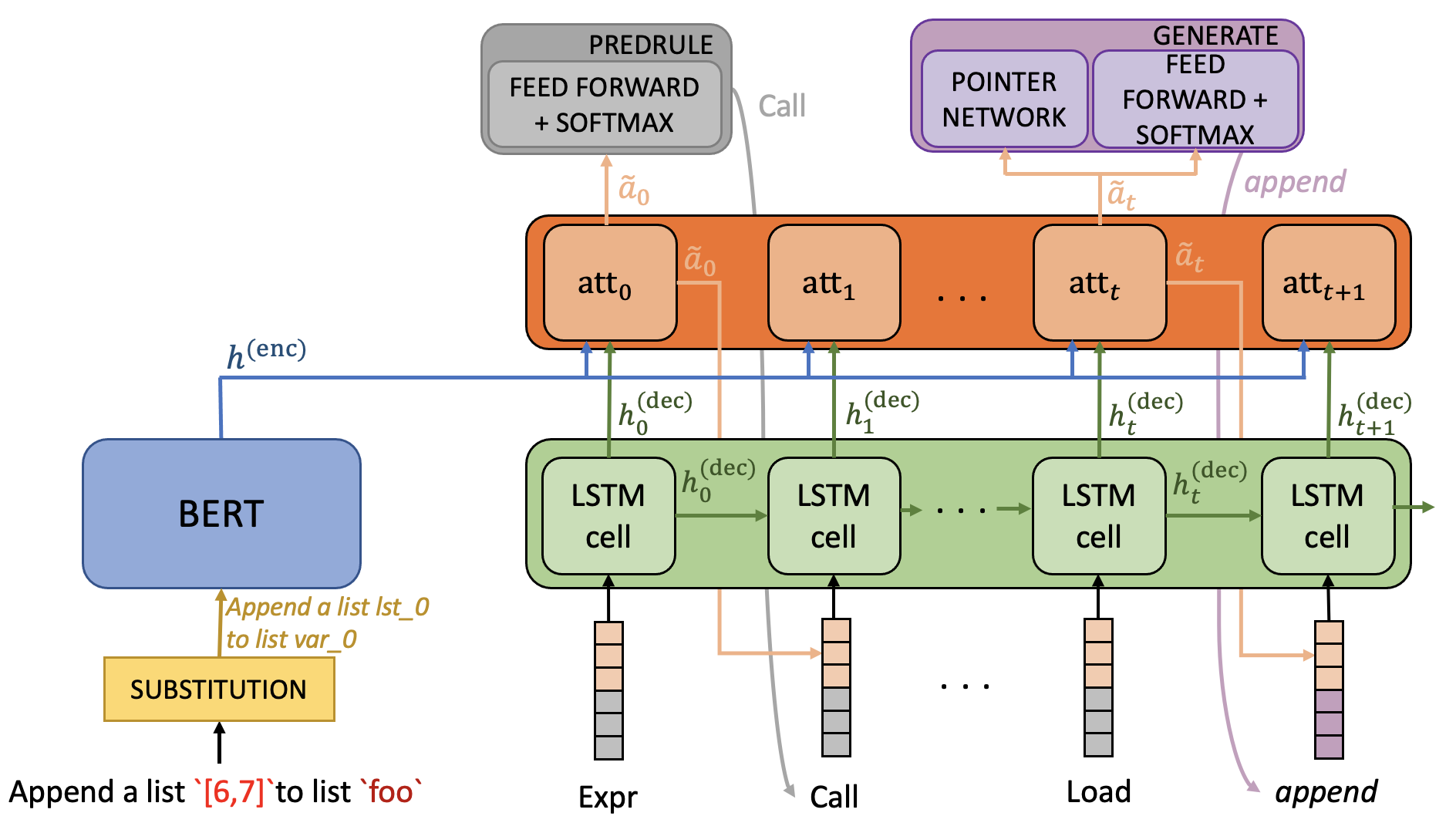}}
\end{center}
\caption{\label{fig:model}Illustration of the seq2seq model with the variables {\var substitution}, {\var grammar}, {\var bert}, {\var pointernet} set to {\var True}. We describe here the complete process where we predict a derivation sequence composed of grammar rules and {\sc close} ({\sc predrule}) or Python variables/built-in ({\sc generate}). The astor library is used to transform the AST constructed with the derivation sequence into Pyton code. In the case where {\var grammar = False}, we only have the {\sc generate} action which exclusively predicts unconstrained code tokens (as for a classical seq2seq).}
\end{figure*}

We consider a family of models generating Python code $y$ from a natural language description $x$, that have the generic form: 
% We employ the transition system from 3.1 to predict a derivation sequence $d$ of grammar rules. Once predicted, this derivation allows the construction of the AST $z$ which ensures the validity of the code:
\begin{equation}
    p({y|x}) = \prod_{t}p({y}_t|y_{<t},x) 
\end{equation}
$y$ is either a sequence of code tokens in case we do not use a grammar, or a sequence of actions from a derivation in case we use a grammar.
The decoding objective aims to find the most-probable hypothesis among all candidate hypotheses by solving the following optimization problem:
% At each decoding step, three types of rules $ d_t$ can be predicted to extend the derivation : \\ \texttt{\textbf{PREDICT}$[c]$}: prediction of a grammar rule from \textit{C} \\ \texttt{\textbf{GENERATE}$[v]$}: prediction of a primitive type. This primitive type can be copied from $x$ or predicted from an external vocabulary $V_{primitives}$. \\ \texttt{\textbf{CLOSE}$[c]$}: closure of a multiple cardinality rule.

\begin{equation}
\label{eqn:MAP}
\hat{y} = \mathop{\text{argmax}}_{y} p(y|x)
\end{equation}

% {\bf TODO : ajustement avec les actions du syst transition}

The family of models varies according to four key qualitative factors that we identify in the {\fontfamily{cmss}\selectfont TranX} architecture. First we describe a substitution procedure managing variables and lists names in section \ref{Preprocessing}). 
Second, in section \ref{subsec:encoder}, we test the architectural variations for encoding the natural language sequence.
Third, in section \ref{subsec:decoder}, we describe variations related to constraining the generated code with grammatical constraints and architectural variations that allow to  copy symbols from the natural language input to the generated code.

\subsection{Substitution} \label{Preprocessing}

Programming languages come with a wide range of variable names and constant identifiers that make the set of lexical symbols infinite.  Rather than learning statistics on a set of ad-hoc symbols, we rather normalize variable and constant names with a pre-processing method, reusing the method of \citet{tranx-2018}.

Preprocessing amounts to substitute the actual names of the variables with a normalized set of predefined names known to the statistical model. The substitution step renames all variables both in the natural language and in the code with conventional names such as \verb+var_0, var_1+, etc. for variables and \verb+lst_0,lst_1+, etc. for lists. A post processing step substitutes back the predicted names with the original variable names in the system output.  For example, given the natural language intent:
\begin{quote}
\textit{create list \texttt{\textasciigrave done\textasciigrave} containing permutations of each element in list \texttt{\textasciigrave [a, b, c, d]\textasciigrave} with variable \texttt{\textasciigrave x\textasciigrave} as tuples}
\end{quote}
is transformed into:
\begin{quote}
\textit{create list \texttt{var\_0} containing permutations of each element in list \texttt{lst\_0} with variable \texttt{var\_1} as tuples}
\end{quote}
% A predicted code such as
%\begin{verbatim}
%var_0 = [(el, var_1) 
%        for el in [lst_0]]
%\end{verbatim}
%is transformed back into 
%\begin{verbatim}
%done = [(el, x) 
%       for el in [a, b, c, d]]
%\end{verbatim}
The predicted code such as \codeword{var_0 = [(el, var_1) for el in [lst_0]]} is transformed back into \codeword{done = [(el, x) for el in [a, b, c, d]]}.

Models using variable replacement as illustrated above, are identified with the notation {\var substitution = True} in section \ref{Experiments}. Implementing this heuristic is made easy by the design of the {\fontfamily{cmss}\selectfont CoNaLa} data set where all such names are explicitly quoted in the data while for {\fontfamily{cmss}\selectfont Django} we had to detect variable names by comparing natural language with its corresponding code.

%define our own heuristic.

% The preprocessing in tranX is based on the setup of the Django and CoNaLa data sets. Both data sets mark variable names with inverted quotes in the natural language. Thus, during preprocessing, all quoted variable names are replaced with \texttt{var\_i} and values of more complex elements such as lists are replaced with \texttt{lst\_i}. This mapping is recorded to allow at inference to replace the predictions of \texttt{var\_i} and \texttt{lst\_i} with the original values and calculate BLEU. \\

%\sout{We adapt this method in Django where most of the variable names are not quoted. The variable names are identified by comparing the natural language with the code and replaced you in the same way as described above. \\
%Table 2 shows the proportion of tokens present in the natural language and in the code. It can be seen that the natural language and the code share a lot of tokens, so that the variable replacement system occurs regularly. This technique leads to a gain of 5.5 points in BLEU for ConaLa and an increase of x points for Django. \\}

%\subsection{Architecture}
%\label{subsec:arch}

\subsection{Encoder}
\label{subsec:encoder}
We switched between a classic bi-LSTM and a pretrained $\mathrm{BERT_{BASE}}$ to encode the input natural language $\{x_{i}, i \in \llbracket 1, n \rrbracket \}$ of $n$ words into a vectorial representations $\{h_{i}^\mathrm{(enc)}, i \in \llbracket 1, n \rrbracket \}$ which are later used to compute the attention mechanism. \\
We set the {\var BERT} factor to {\var True} when using it and {\var False} when using the bi-{\sc lstm}.
% \sout{To match the size of the decoder's hidden layer and calculate the attention weights, we apply a feed-forward layer on the hidden vector $\hat{h}$.}

\subsection{Decoder}
\label{subsec:decoder}

At each time step $t$, the $\mathrm{LSTM}$ decoder computes its internal hidden state $h_{t}^\mathrm{(dec)}$: 

\begin{equation}
{ h_{t}^\mathrm{(dec)}} = \text{LSTM}([e_{t-1}:\widetilde{a}_{t-1}], h_{t-1}^\mathrm{(dec)})
\end{equation}
where $e_{t-1}$ is the embedding from the previous prediction, $\widetilde{a}_{t-1}$ is the attentional vector. 

We compute the attentional vector $\widetilde{a}_{t}$ as in \citet{Attention} combining the weighted average over all the source hidden state $c_t$ and the decoder hidden state $h_{t}^\mathrm{(dec)}$: 
\begin{equation}
\widetilde{a}_t = W_a[c_t:h_{t}^\mathrm{(dec)}]
\end{equation}
It is the attention vector $\widetilde{a}_{t}$ which is the key to determine the next prediction $y_t$.

%Different types of prediction are presented depending on the variables {\var pointernet} and {\var grammar}. \\
%In the case where code tokens are predicted i.e. {\var grammar} = {\bf False}, there are two possibilities depending on the use of the copy mechanism.
%On the one hand, when code tokens are predicted i.e. {\var grammar} = {\bf False}, there are two possibilities depending on the use of the copy mechanism.

We use several variants of the code generator, that we describe by order of increasing complexity. The  basic generator is a feed forward that uses the attention vector to generate a code token $v$ from a vocabulary $V$:
\begin{equation}
\label{eqn:generate}
\begin{aligned}
    p(y_t = \mathrm{GENERATE}[v]|x, e_{<t}) = \\ 
    \text{softmax}(e_v^\top \cdot W_g \cdot {\widetilde{a}_t})
\end{aligned}
\end{equation}
These models are not constrained by the Python grammar and we identify these models with {\var grammar = False}. 

We also use a pointer network that may either copy symbols from input to output or generate symbols from $V$. Then the probability of generating the symbol $v$ is given by the marginal probability:
\begin{equation}
\label{eqn:generate+pn}
\begin{aligned}
    p(y_t = \mathrm{GENERATE}[v]|x, e_{<t})= \\ 
    p(\mathrm{gen}|x,e_{<t}) p(v|\mathrm{gen}, x, e_{<t}) \\
    + p(\mathrm{copy}|x,e_{<t}) p(v|\mathrm{copy}, x, e_{<t})
\end{aligned}
\end{equation} 
The probabilities $p(\mathrm{gen}|.)$ and $p(\mathrm{copy}|.)$ sum to 1 and are computed with $\text{softmax}(W \cdot \widetilde{a}_t)$. The probability of generating $v$ from the vocabulary $V$ $p(v|\mathrm{gen}, .)$ is defined in the same way as (\ref{eqn:generate}). We use the pointer net architecture \cite{Pointernet} to compute the probability $p(v|\mathrm{copy},.)$ of copying an element from the natural language $x$. Models that use a pointer network are identified with {\var pn = True}, otherwise with {\var pn = False} . 

Finally we use a set of models that are constrained by the Python grammar and that rely on the transition system from section \ref{TransitionSystem}. Rather than directly generating Python code, these models generate a derivation whose actions are predicted using two prediction tasks. \\ 
When the generator is in a state where the dot of the item on the top of the stack points on a nonterminal symbol, the {\sc predrule} is used. This task either outputs a {\sc predict}(C) action or the {\sc close} action:
\begin{equation}
\label{eqn:applyrule}
\begin{aligned}
    p(y_t = \mathrm{PREDRULE}[c]|x, e_{<t})= \\ 
    \text{softmax}(e_r^\top \cdot W_p \cdot {\widetilde{a}_t})
\end{aligned}
\end{equation}
When the generator is in a state where the dot of the item on the top of the stack points on a terminal symbol, the generate task is used.
This amounts to reuse either equation (\ref{eqn:generate}) or equation (\ref{eqn:generate+pn}) according to the model at hand. Models constrained by the grammar are labelled with  {\var grammar = True}. 
Recall that the {\sc complete} action of the transition system is called deterministically (Section \ref{TransitionSystem}).

% \texttt{\textbf{PREDICT}$[c]$}: prediction of a grammar rule from \textit{C} \\ \texttt{\textbf{GENERATE}$[v]$}: prediction of a primitive type. This primitive type can be copied from $x$ or predicted from an external vocabulary $V_{primitives}$. \\ \texttt{\textbf{CLOSE}$[c]$}: closure of a multiple cardinality rule.

%\subsection{Difference with TranX}
%\begin{itemize}
%  \item AST is not constructed at each inference step's
%\end{itemize}
% $W_g$ where each row of $W_r$ (respectively $W_g$) is an embedding vector for \texttt{PREDICT} (respectively \texttt{GENERATE})

\comment{
\subsubsection{Action Probabilities}

{\sl comment on grammatical constraints;  pointer network }

The attentional vector $\widetilde{s}_t$ is the key to compute the next derivation's action. \\
\textbf{\texttt{PREDICT} case} The probability of using rule r as the current action $d_t$ is given by softmax:
\begin{equation}
\label{eqn:applyrule}
\begin{aligned}
    p(d_t = PREDICT[r]|x, d_{<t})= \\ 
    \text{softmax}(a_t^T \cdot W \cdot {(\widetilde{s}_t})) . e(r)
\end{aligned}
\end{equation}

where each row of $a_t$ is embedding matrix for \texttt{PREDICT} and e(r) is the one-hot vector for rule r. \\
\textbf{\texttt{GENERATE} case} As explained in \nameref{Code Generation}, a new token $v$ can be generated from an external vocabulary or copied from the input. Then the probability of generating $v$ is given by the marginal probability:
\begin{equation}
\begin{aligned}
    p(d_t = GENERATE[v]|x, d_{<t})= \\ 
    p(gen|x,d_{<t}) p(v|gen, x, d_{<t}) \\
    + p(copy|x,d_{<t}) p(v|copy, x, d_{<t}) 
\end{aligned}
\end{equation}
{\sl distinguer le cas où on utilise le PN et la cas où on l'utilise pas}

Here, the probabilities p(gen|.) and p(copy|.) are computed with $\text{softmax}(W . \widetilde{s}_t)$. The probability of generating v from an external vocabulary $p(v|gen, .)$ is defined in the same way as \ref{eqn:applyrule} (with $W_g$ as embedding matrix for \texttt{GENERATE}). We use the pointer net architecture \cite{Pointernet} to compute the probability $p(v|copy,.)$ of copying an element from the natural language $x$.}

\section{Experiments} 
\label{Experiments}

In this section we describe the characteristics of the data sets on which we have tested our different setups and the underlying experimental parameters\footnote{The code of our experiments is public and available at {\tt https://gitlab.com/codegenfact/BertranX}}. %https://github.com/Jonor127-OP/NL2CODE\_baseline}

\comment{
\subsection{Setup}

{\bf Data sets} In this study we use two available data sets, {\fontfamily{cmss}\selectfont Django} and {\fontfamily{cmss}\selectfont CoNaLa}, to perform our code generation task. 

The {\fontfamily{cmss}\selectfont Django} data set provides line-by-line comments with code from the {\fontfamily{cmss}\selectfont Django} web framework. About 70\% of the 18805 examples are simple Python operation ranging from function declarations to package imports, and including exception handling.
Those examples strongly share the natural language structure (e.g. {\it call the function cache.close} $\rightarrow$ \texttt{cache.close()}) with more than 26\% of the words in the natural language are also present in the code. The BLEU score between the NL-code is equal to 19.4.
% This data set contains short code descriptions mapped to short python one-liners. Table \ref{tab-lex-desc} further shows that the comments and the code share some structure in this data set as can be seen from the relatively high BLEU score we get by simply comparing literally the reference input and output.

{\fontfamily{cmss}\selectfont CoNaLa} is made up of 600k natural language-code pairs from {\tt StackOverflow}, part of which have been manually cleaned up by developers (2879 examples). The natural language descriptions correspond to actual developer queries (e.g. {\it Delete an element 0 from a dictionary `a`}) and the associated code is diverse and idiomatic ({e.g. \tt \{i: a[i] for i in a if (i != 0)\}}). Compared to {\fontfamily{cmss}\selectfont Django}, the code is therefore much more difficult to generate. In particular because the number of words shared between the NL and the code is much lower (BLEU = 0.32). Also, the code is longer and more complex with an AST depth of 7.1 on average against 5.1 for {\fontfamily{cmss}\selectfont Django}.

{\bf Configuration} Hyperparameters of our model depend on whether BERT is used or not. \\
When {\var BERT = False}, the size of all embeddings is 128. The hidden layers size of the bi-LSTM at encoding and the LSTM at decoding is 256. The resulting attention vector size is 300. Since our data sets are relatively small for a data-hungry neural model, we impose regularization  with a word dropout layer for embeddings and dropout at the output of the attention.\\
When {\var BERT = True}, the size of the natural language embedding is already set to 756 with BERT.  We therefore apply a linear transformation to its output so that the embedding size is equal to 512.The embedding of the code tokens is fixed at 256, as well as the embeddings of the grammar actions when {\var grammar = True}. LSTM decoder hidden state and attention vector are set at 512. We regularize only the attentional vector in that case.\\
In both case, we use a beam search size of 15 for decoding. \\}

\subsection{Data sets}

In this study we use two available data sets, {\fontfamily{cmss}\selectfont Django} and {\fontfamily{cmss}\selectfont CoNaLa}, to perform our code generation task. 

The {\fontfamily{cmss}\selectfont Django} data set provides line-by-line comments with code from the {\fontfamily{cmss}\selectfont Django} web framework. About 70\% of the 18805 examples are simple Python operation ranging from function declarations to package imports, and including exception handling.
Those examples strongly share the natural language structure (e.g. {\it call the function cache.close} $\rightarrow$ \texttt{cache.close()}). More than 26\% of the words in the natural language are also present in the code, BLEU score between the natural language and code is equal to 19.4.
% This data set contains short code descriptions mapped to short python one-liners. Table \ref{tab-lex-desc} further shows that the comments and the code share some structure in this data set as can be seen from the relatively high BLEU score we get by simply comparing literally the reference input and output.

{\fontfamily{cmss}\selectfont CoNaLa} is made up of 600k NL-code pairs from {\tt StackOverflow}, among which 2879 examples have been been manually cleaned up by developers. %\footnote{Actual number is lower due to errors in the data set preventing the usage of some examples}. 
All results are reported on the manually curated examples, unless stated otherwise. The natural language descriptions are actual developer queries (e.g. {\it Delete an element 0 from a dictionary `a`}) and the associated code is diverse and idiomatic ({e.g. \tt \{i: a[i] for i in a if (i != 0)\}}). Compared to {\fontfamily{cmss}\selectfont Django}, the code is much more challenging to generate. Especially because the number of words shared between the NL and the code is much lower (BLEU = 0.32). Also, the code is longer and more complex with an AST depth of 7.1 on average against 5.1 for {\fontfamily{cmss}\selectfont Django}.

% \textbf{TODO : (SEE APPENDIX FOR EXAMPLES).}

\subsection{Vocabulary generation}

The vocabulary of natural language and code is essential. Usually, this vocabulary is created by adding all the words present in the training data set. There are however exceptions that are detailed in this section.

The natural language vocabulary relies on a byte pair encoding tokenizer when {\var BERT = True}. As explained in section \ref{Preprocessing}, the variable names are replaced with special tokens \texttt{var\_i} and \texttt{lst\_i}. These new tokens are crucial to our problem, and added to the BERT vocabulary %\footnote{\textcolor{red}{We add new tokens with the function \texttt{add\_tokens} from \texttt{transformers} library}}
. We can then fine-tune BERT with this augmented vocabulary on our data sets.

For the decoder part, when {\var grammar = True}, the vocabulary of grammatical actions is fixed, while the vocabulary of AST leaves has to be built. This associated vocabulary can be composed of built-in Python functions, libraries with their associated functions or variable names. Its creation is consequently a major milestone in the generation process. 

To create this external vocabulary, we proceed as in {\fontfamily{cmss}\selectfont TranX}. From the code, we create the derivation sequence composed of the action of the grammar as well as the primitives. All primitives of the action sequences are incorporated into our external vocabulary. 
%One could imagine a more general external vocabulary less focused on  examples with Python built-in functions and specialized by libraries (assuming that variable names are always reported by quotes).

\subsection{Setup}

%{\bf Configuration} There are at least two embeddings, for natural language and code tokens. When using grammar, there is embedding for grammar rules as well. Since our data sets are relatively small for a data-hungry neural model, we choose short embedding size and impose strong regularization. \\
% Hyperparameters of our model depend on whether BERT is used or not. \\
When {\var BERT = False}, the size of the representations is kept small to prevent overfitting. Encoder and decoder embedding size is set to 128. The hidden layer size of the encoder and decoder bi-LSTM is set to 256 and the resulting attention vector size is 300.   
We have two dropout layers: for embeddings and at the output of the attention.  We use Adam optimizer with learning rate $\alpha = 5.10^{-3}$.

When {\var BERT = True}, encoder embeddings have a natural size of 756 with BERT. We therefore apply a linear transformation to its output to get an embedding size equal to 512. The size of LSTM decoder hidden state and attention vector are set to 512. We regularize only the attentional vector in that case. We use Adam optimizer with learning rate $\alpha = 5.10^{-5}$. In both cases, we use a beam search size of 15 for decoding. 

%{\bf Validation set} We removed 10\% examples from the training
%set to form the validation set on both data set. \\
\paragraph{Evaluation} To compare with previous work, we report the standard evaluation metrics for each data set: exact match accuracy and corpus-level BLEU. %\textcolor{red}{We choose these two evaluation metrics to compare with previous work.} \\

\paragraph{Python version} As the grammar slightly changes between Python versions, let us mention that all our experiments have been carried out with Python 3.7.

\subsection{Ablation study} 
\label{Ablation}

%\subsection{Results}
%\label{sec:results}

\begin{table*}[htbp]
\begin{center}
\scalebox{0.85}{
\begin{tabular}{ |c|c|c|c|c|c|c|c| } 
\hline
Substitution & {\sc Bert} & Grammar & PN & {\fontfamily{cmss}\selectfont CoNaLa} BLEU & {\fontfamily{cmss}\selectfont CoNaLa} accuracy & {\fontfamily{cmss}\selectfont Django} BLEU & {\fontfamily{cmss}\selectfont Django} accuracy  \\
\hline
\multirow{8}{2em}{False} & \multirow{4}{2em}{False} & \multirow{2}{2em}{False} & \multirow{1}{2em}{False} & $21.05 \pm 0.81$    & $0.9\pm 0.42$  & $42.58 \pm 1.54$ & $26.86 \pm 1.15$  \\ \cline{4-8}
&  &  & \multirow{1}{2em}{True} &  $22.33 \pm 0.78$    & $1.7\pm 0.90$  & $64.79 \pm 1.00$ & $62.85 \pm 1.21$  \\ \cline{3-8}
&  & \multirow{2}{2em}{True} & \multirow{1}{2em}{False} & $20.59 \pm 0.74$    & $2.87 \pm 0.48$  & $43.23 \pm 1.62$ & $30.12\pm 0.63$   \\ \cline{4-8}
&  &  & \multirow{1}{2em}{True} &$22.16 \pm 1.93$     & $3.87 \pm 1.65$ & $62.55 \pm 1.60$ & $65.20 \pm 0.03$    \\ \cline{2-8}
& \multirow{4}{2em}{True} & \multirow{2}{2em}{False} & \multirow{1}{2em}{False} &  $30.83\pm 4.08$ & $2 \pm 0.94$ & $53.18 \pm 0.87$ & $30.28 \pm 0.26$ \\ \cline{4-8}
&  &  & \multirow{1}{2em}{True} & $30.98 \pm 1.33$ & $3.3 \pm 1.48$ & $58.69 \pm 1.28$ & $37.96 \pm 0.27$   \\ \cline{3-8}
&  & \multirow{2}{2em}{True} & \multirow{1}{2em}{False} & $25.88 \pm 0.94$  & $3.8 \pm 1.96$ & $47.32 \pm 0.50$ & $29.62 \pm 0.33$ \\ \cline{4-8}
& \multirow{6}{2em}{False}  &  & \multirow{1}{2em}{True} & $28.43 \pm 0.64 $& $4.4 \pm 1.67$ & $52.55 \pm 0.51$ & $37.38 \pm 0.38$   \\ \cline{1-8} 
&  & \multirow{2}{2em}{False} & \multirow{1}{2em}{False} & $31.17 \pm 0.88$    & $3.1\pm 1.52$  & $70.4 \pm 0.25$ & $70.40 \pm 0.29$   \\ \cline{4-8} 
\multirow{10}{2em}{True} &  &  & \multirow{1}{2em}{True} & $32.10 \pm 1.06$  & $3.1 \pm 1.24$& $70.28 \pm 0.38$& $70.46 \pm 0.37$  \\ \cline{3-8}
&  & \multirow{2}{2em}{True} & \multirow{1}{2em}{False} & $33.36 \pm 1.63$    & $6.37 \pm 0.63$  & $70.82 \pm 0.22$ & $71.3 \pm 0.19$ \\ \cline{4-8}
&   &  & \multirow{1}{2em}{True}  & $32.86 \pm 1.75$  & $5 \pm 1.67$& $70.62 \pm 0.49$& $71.47 \pm 0.19$   \\  \cline{2-8}
& \multirow{6}{2em}{True} & \multirow{3}{2em}{False} & \multirow{1}{2em}{False} & $36.43\pm 0.41$ & $4.5 \pm 1.84$ & $\boldsymbol{76.97 \pm 0.15}$ & $74.58 \pm 0.27$  \\ \cline{4-8} 
&  &  & \multirow{2}{2em}{True} & $36.29 \pm 2.27$ & $5 \pm 1.32$ &$76.62 \pm 0.50$& $\boldsymbol{76 \pm 0.71}$  \\ 
&  &  &  & $35.42 \pm 1.75^*$ & $5.2 \pm 1.33^*$ & - & - \\ \cline{3-8} 
&  & \multirow{3}{2em}{True} & \multirow{1}{2em}{False} & $35.04 \pm 1.03$ & $7.3 \pm 1.25$ & $76.20 \pm 0.46$ & $74.88 \pm 0.56$ \\ \cline{4-8}
&  &  & \multirow{2}{2em}{True} & $37.99 \pm 1.85$ & $7.5 \pm 1.12$ &$76.32 \pm 0.59$& $75.32 \pm 1.54$  \\ 
&  &  &  & $\boldsymbol{39.01 \pm 1.08}^*$ & $\boldsymbol{ 7.7 \pm 1.92}^*$ & - & -  \\ 
\hline
\end{tabular}}
\caption{Performances with different natural language encoders on the development sets with and without a grammatical component. The scores reported are the mean and standard deviation resulting from training with 5 different seeds. The * refers to the use of 100k {\fontfamily{cmss}\selectfont CoNaLa} mined data in addition to clean examples.}
\label{table:ablationstudy}
\end{center}
\end{table*}

\begin{figure}[htbp]
\scalebox{0.48}{

\includegraphics[width=\textwidth]{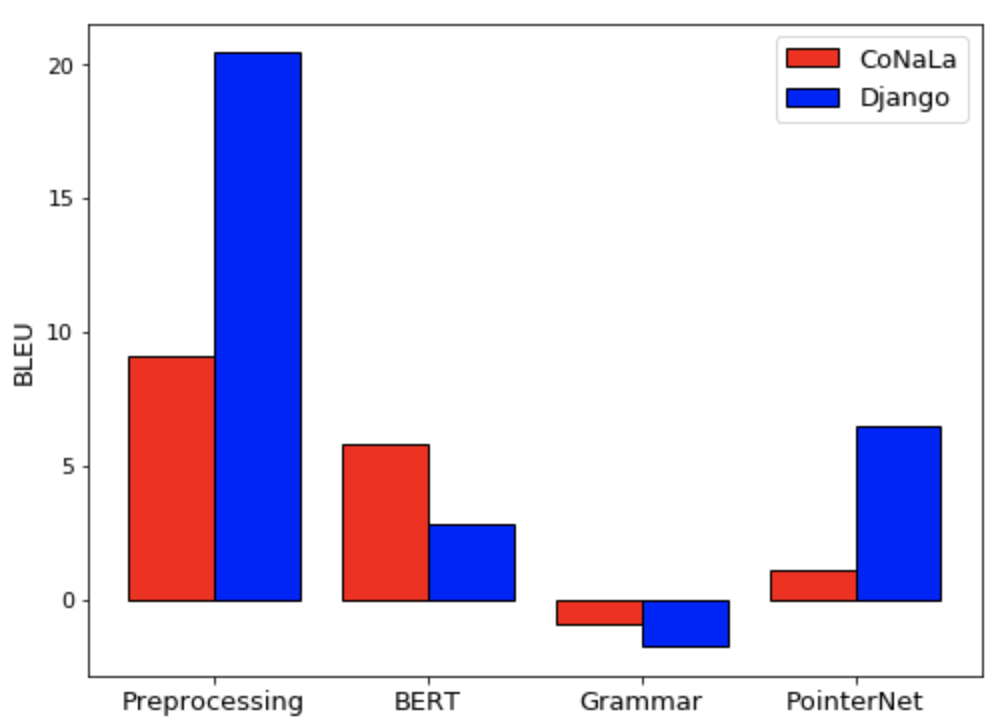}}
\caption{Difference between the marginal mean of each variable for the {\var True} and {\var False} conditions. }
\label{fig:margmean}
\end{figure}
% Next, we report state of the art result on both data sets thanks to our best model in Section \ref{testset}.

To  highlight the contribution of the different factors, {\sc substitution, bert, grammar, pn} on the  {\fontfamily{cmss}\selectfont Django} and {\fontfamily{cmss}\selectfont CoNaLa} data sets
we report a detailed study of their impact in Table \ref{table:ablationstudy}.

The results are analyzed by distinguishing lexical and grammatical aspects and by identifying relations between the different factors. We start by a comparison of the marginal mean of the BLEU score for each of our variables in both conditions. Figure \ref{fig:margmean} highlights the mean difference between the conditions by contrasting the case where the value is {\sc True} with the case where the value is {\sc False}.

\comment{
\begin{figure}[htbp]
\begin{tikzpicture}
\begin{axis}[
    symbolic x coords={substitution, BERT, grammar, pointernet},
    xtick=data,ylabel=BLEU, ybar,height=7cm,width=8cm]
    \addplot[ybar,fill=blue] coordinates {
        (substitution,9.13)
        (BERT,5.78)
        (grammar,-0.91)
        (pointernet,1.1)
    };
    \addplot[ybar,fill=red] coordinates {
        (substitution,20.41)
        (BERT,2.83)
        (grammar,-1.73)
        (pointernet,6.47)
    };
\legend{CoNaLa, Django}
\end{axis}
\end{tikzpicture}
\caption{Difference between the marginal mean of each variables set to True and False. {\it la légende est dégueu je sais pas comment tourner ça}}
\label{fig:margmean}
\end{figure}
\\}

\begin{table*}[htbp]
\scalebox{0.8}{
\begin{tabular}{l|cccc}
 \hline
System & {\fontfamily{cmss}\selectfont CoNaLa} BLEU & {\fontfamily{cmss}\selectfont CoNaLa} accuracy & {\fontfamily{cmss}\selectfont Django} BLEU & {\fontfamily{cmss}\selectfont Django} accuracy \\
\midrule
\citep{tranx-2018}  & 27.2 & - & - & 73.7\\ 
\citep{tranx-2018} + mined  & 28.1    & -  & - & - \\
\citep{orlanski-21} + mined 100k &30.55&-&-&-\\
\citep{monolingualdata} + 600k mined & 32.57&-&-& {\bf 81.03}\\
\midrule
% Ours {\sc Lstm + grammar}  &28.5   & 2.7& 73 & 74.79 \\ 
Ours {\sc Bert + grammar}  & 31.6   & 4.5 & 79.86 & 79.77 \\ 
Ours {\sc bert + grammar} + 100k mined  &  {\bf 34.20} & {\bf 5.8} & - & - \\ \hline
% Ours {\sc Lstm } (tokens) & 29.82  & 2 & 74.36 & 74.02 \\ 
Ours {\sc Bert} (tokens) & 30.73   & 1.40 & 79.81 & 79.61 \\ 
Ours {\sc bert} + 100k mined (tokens) &  32.39 & 3.4 & - & - \\ \hline
\end{tabular}}
\caption{Comparisons of the systems trained without external data sources on {\fontfamily{cmss}\selectfont CoNaLa} and {\fontfamily{cmss}\selectfont Django} test sets.}
\label{table:sota}
\end{table*}
% We observe the lexical importance through preprocessing and pointer network.

\paragraph{Pointer network} The pointer network can improve the results, especially when {\var substitution = False}. This is because the only way to obtain the name of the variables is to copy them. 
% The results with BERT drop because the byte pair tokenizer slices the variables (66\% are sliced for {\fontfamily{cmss}\selectfont CoNaLa}, e.g. \codeword{mylist} becomes \codeword{my} and \codeword{##list}). 
%When there is preprocessing, there is no need to predict variable names directly because we integrated \codeword{var_i, lst_i} into the external vocabulary. 
Combined with substitution, the pointer network offers an additional possibility to predict the \codeword{var_i, lst_i} which allows to achieve the best results with a BLEU score of 39.01 on {\fontfamily{cmss}\selectfont CoNaLa} and an exact match accuracy of 76 on {\fontfamily{cmss}\selectfont Django}. 
\paragraph{Substitution and Typing} The scores are stabilised and much higher with substitution. We gain more than 9 points of BLEU on {\fontfamily{cmss}\selectfont CoNaLa} (respectively 20 points on {\fontfamily{cmss}\selectfont Django}) thanks to substitution.  The "weakest" configuration where all variables are {\sc False} except the substitution gives better results than all configurations where {\var substitution = False}. \\
The increase in BLEU with substitution can be explained in two ways.
On the one hand, we remark that the model has difficulties to memorize the values to fill the lists with {\sc generate}. For example, four tokens of code must be generated to predict the list \codeword{[a, b, c, d]}. Using substitution, the model can just predict \texttt{lst\_0} which will be replaced by \codeword{[a, b, c, d]} during postprocessing. This avoids a potential error in the creation of the list and directly gives a valid 4-gram. This contributes to greatly increase the BLEU, which shows the importance of replacing lists. On {\fontfamily{cmss}\selectfont CoNaLa}, BLEU score on the development set drops from an average of 37.99 to an average of 30.66 without list replacement. 
%The jump to BLEU with substitution can be explained by the replacement of lists which drastically increases the count of valid n\_grams. For example, just by predicting \texttt{lst\_0} instead of \codeword{['a', 'b', 'c', 'd']}, we avoid predicting 4 variables separately and we greatly increase the BLEU because multiple n\_grams are provided. {\bf TODO: ajouter résultats score sans listes} \\
Besides list replacement, the architecture has also a weakness with respect to variable typing. 
When using the grammar without substitution, the results are lower than without grammar.
This effect is the result of a type checking failure. The model predicts ill-typed AST structures. For instance it predicts an AST whose corresponding code should be \codeword{1.append([6,7])}. However the AST library we used prevents from generating such ill-typed code. The absence of code generation in such cases explain the decrease in BLEU score.

The use of substitution partially corrects for these typing errors because the substituted symbols \codeword{var_i, lst_i} are generally more likely to be predicted and are likely to have the right type thanks to the mapping.

{\bf Grammatical aspect} The transition system doesn't improve the results on average because of the empty predictions when {\var substitution = False}. The use of the transition system leads to better results when {\var substitution = True} but not as drastically as one would have expected. 
However the real contribution of the grammar associated with substitution is the syntactic validity of the code in 100\% of the cases, as tested with our architecture obtaining the best results. In scenarios where we do not use the grammar, it is never the case to have an empty output. But then the proportion of code sequences that are actually syntactically valid in this setup is 92\% on average. 

% The jump in score is mainly due to the use of BERT and preprocessing. 

\paragraph{BERT} As expected when using BERT to encode the natural language input we get an improvement of about 6 marginal BLEU on {\fontfamily{cmss}\selectfont CoNaLa} (respectively +3 BLEU on {\fontfamily{cmss}\selectfont Django}). More interestingly, this effect is lower than the one of the substitution operation.
% We notice that the results seem to have a ceiling on Django, especially when BERT is used with preprocessing with or without grammar. As the Python code snippets to be generated are very simple and close to natural language, the similarity between the results with and without grammar seems logical.  Django's Python code has many inaccurate examples that the network will not be able to reconstruct which may explain the ceiling, more preprocessing should probably be done. 
\paragraph{}We conclude that the use of a pre-trained model increases the results but less than substitution, despite what one might think and it suggests that improving the management of variable names and lists is one of the key elements for improving the system. The contribution of grammatical constraints in BLEU may seem detrimental but we could see that this is a side effect of typing constraints in adversarial scenarios. Overall the non-constrained generated code is syntactically incorrect in 8\% of the cases.

%For CoNaLa, there is not necessarily a ceiling but rather a certain regularity with the use of the grammar. In particular, the grammar stabilises the results when using the mined examples in addition to the manually curated examples.
%The first striking thing concerns the use of the transition system, which gives better results than code token prediction, but not as drastically as one would have expected. 

%The use of BERT without preprocessing gives very bad results especially for Django because its specific vocabulary is much richer than CoNaLa's one, so the variable names are often replaced by \texttt{UNK} by the BERT tokenizer. 

%It is noticeable that the results seem to be capped on Django when using BERT with preprocessing with or without grammar and pointer network. As the Python code snippets to be generated are very simple and close to natural language, the similarity between the results with and without grammar seems logical.  Django's Python code has many inaccurate examples that the network will not be able to reconstruct which may explain the ceiling, more preprocessing should probably be done. 

\comment{
\begin{table*}
\scalebox{0.85}{
\begin{tabular}{cccc|cc |cc}
 \toprule
 {\sc Bert} & grammar & preprocessing &PN&CoNaLa BLEU & CoNaLa accuracy & Django BLEU & Django accuracy \\ \midrule

    no&no&no&no& $21.05 \pm 0.81$    & $0.9\pm 0.42$  & $42.58 \pm 1.54$ & $26.86 \pm 1.15$ \\
  yes &no&no&no  &  $30.83\pm 4.08$ & $2 \pm 0.94$ & $53.18 \pm 0.87$ & $30.28 \pm 0.26$ \\  
  \midrule
  
   no &no&no&yes& $22.33 \pm 0.78$    & $1.7\pm 0.90$  & $64.79 \pm 1.00$ & $62.85 \pm 1.21$ \\
  yes &no&no&yes  &  $30.98 \pm 1.33$ & $3.3 \pm 1.48$ & $58.69 \pm 1.28$ & $37.96 \pm 0.27$ \\  
  \midrule
 no &no&yes&no& $31.17 \pm 0.88$    & $3.1\pm 1.52$  & $70.4 \pm 0.25$ & $70.40 \pm 0.29$ \\
  yes &no&yes&no  &  $36.43\pm 0.41$ & $4.5 \pm 1.84$ & $76.97 \pm 0.15$ & $74.58 \pm 0.27$ \\  
  \midrule
  no &no &yes&yes&$32.10 \pm 1.06$  & $3.1 \pm 1.24$& $70.28 \pm 0.38$& $70.46 \pm 0.37$ \\ 
  yes & no &yes &yes& $36.29 \pm 2.27$ & $5 \pm 1.32$ &$76.62 \pm 0.50$& $76 \pm 0.71$ \\ 
 yes* & no &yes&yes & $35.42 \pm 1.75$ & $5.2 \pm 1.33$ & - & - \\ \midrule
   no  &yes&no&no&$20.59 \pm 0.74$    & $2.87 \pm 0.48$  & $43.23 \pm 1.62$ & $30.12\pm 0.63$ \\
 yes &yes&no&no & $25.88 \pm 0.94$  & $3.8 \pm 1.96$ & $47.32 \pm 0.50$ & $29.62 \pm 0.33$ \\ \midrule
 no &yes&no&yes& $22.16 \pm 1.93$     & $3.87 \pm 1.65$ & $62.55 \pm 1.60$ & $65.20 \pm 0.03$ \\
 yes&yes &no&yes  & $28.43 \pm 0.64 $& $4.4 \pm 1.67$ & $52.55 \pm 0.51$ & $37.38 \pm 0.38$ \\ \midrule
  no &yes&yes&no& $33.36 \pm 1.63$    & $6.37 \pm 0.63$  & $70.82 \pm 0.22$ & $71.3 \pm 0.19$ \\
 yes &yes&yes&no  &  $35.04 \pm 1.03$ & $7.3 \pm 1.25$ & $76.20 \pm 0.46$ & $74.88 \pm 0.56$ \\ 
 \midrule
 no &yes &yes&yes&$32.86 \pm 1.75$  & $5 \pm 1.67$& $70.62 \pm 0.49$& $71.47 \pm 0.19$ \\ 
 yes & yes &yes &yes& $37.99 \pm 1.85$ & $7.5 \pm 1.12$ &$76.32 \pm 0.59$& $75.32 \pm 1.54$ \\ 
 yes* & yes &yes&yes & $39.01 \pm 1.08$ & $7.7 \pm 1.92$ & - & - \\ \bottomrule
\end{tabular}}
\caption{Performances with different natural language encoders on the development sets with and without a grammatical component. The scores reported are the mean and standard deviation resulting from training with 5 different seeds. The * refers to the use of CoNaLa mined data in addition to clean examples.}
\label{table:ablationstudy}
\end{table*}
}

\subsection{Test}
\label{testset}

We compare in table \ref{table:sota} our results with other systems on {\fontfamily{cmss}\selectfont CoNaLa} and {\fontfamily{cmss}\selectfont Django} test sets. 
We report our best performing models on the development set with and without grammatical constraints. We also use models trained on the full {\fontfamily{cmss}\selectfont CoNaLa} including mined examples to get relevant comparisons.

Among the other systems \citet{tranx-2018} is the only one that uses grammatical constraints. Our architecture differs with the use of a BERT  encoder whereas \citet{tranx-2018} use an LSTM.
The other systems do not use grammatical constraints but rather try to take advantage of additional data.  \citet{orlanski-21} and \citet{monolingualdata} aim to take advantage of the {\fontfamily{cmss}\selectfont CoNaLa} mined examples. As these mined examples are noisy, \citet{orlanski-21} takes advantage of BART \cite{BART}, a denoising encoder. They also enrich the natural language input with the results of queries from StackOverflow by adding the title of the post, its associated tags, etc.
\citet{monolingualdata} use  BERT as encoder and a transformer decoder. They apply the Target Autoencoding method introduced by \citet{TAE}. During training, the encoder parameters are frozen and the decoder is trained to reconstruct code examples. They use this method on the mined examples to take maximal advantage of the additional noisy data.

We observe that our grammar based model with BERT encoder is state of the art on {\fontfamily{cmss}\selectfont CoNaLa} while the transformer encoder/decoder architecture of \citet{monolingualdata} performs best on {\fontfamily{cmss}\selectfont Django}. Quite interestingly the exact match accurracy of these models remain weak on {\fontfamily{cmss}\selectfont CoNaLa}.

%With grammar, our BERT based model yield to better results on {\fontfamily{cmss}\selectfont CoNaLa} (BLEU score of 34.20 against 32.57 for \citet{monolingualdata}) and  equivalent results on {\fontfamily{cmss}\selectfont Django} (accuracy of 79.77 against 81.03 for \citet{monolingualdata}).

%We present the results obtained with and without grammar on the {\fontfamily{cmss}\selectfont CoNaLa} and {\fontfamily{cmss}\selectfont Django} test sets on Table \ref{table:sota}. Results obtained are superior or competitive with state-of-the-art models specifically designed for semantic parsing.  \\

\section{Conclusion}

We formalized a transition system that allows us to guarantee the generation of syntactically correct code. A detailed study of the components of the seq2seq architecture reveals that the models have difficulties at managing accurately variable names and list encodings. The comparison with models trained on larger noisy data sets reveals that our grammatically constrained architecture without explicit denoising remains competitive. This further highlights the importance of grammatical constraints and of specific processes dedicated to manage variables, list naming and typing. 

Finally, we observe that BLEU and exact match, used in this paper, although commonly used in the literature, are not ideal metrics especially
because high BLEU scores do not guarantee that the code will be executable. Even exact match is not satifactory since a single natural language query can be solved by several python programs.
In future work, we plan to 
 build extensions to the datasets used here with additional test cases assessing the correction of the generated code.
These tests are likely to support more relevant metrics for  code generation evaluation. 

%We propose a new transition system that allows us to compare different factors on Python code generation. We studied the impact of lexical factors such as substitution and pointer network, the importance of natural language with BERT, as well as syntactic contribution using the grammar. Although the contribution of a pre-trained model such as BERT increases the scores, we show that substitution plays a central role in Python code prediction. This substitution allows in part to ensure the validity of the typing while reducing the possibility space of the variables to be predicted. Combined with the grammar, the results are state of the art with a strong assurance that the generated code is compilable.

% Entries for the entire Anthology, followed by custom entries

\bibliography{anthology,custom}

\begin{thebibliography}{18}
\expandafter\ifx\csname natexlab\endcsname\relax\def\natexlab#1{#1}\fi

\bibitem[{Austin et~al.(2021)Austin, Odena, Nye, Bosma, Michalewski, Dohan,
  Jiang, Cai, Terry, Le, and Sutton}]{LLM1}
Jacob Austin, Augustus Odena, Maxwell Nye, Maarten Bosma, Henryk Michalewski,
  David Dohan, Ellen Jiang, Carrie~J. Cai, Michael Terry, Quoc~V. Le, and
  Charles Sutton. 2021.
\newblock \href {http://arxiv.org/abs/2108.07732} {Program synthesis with large
  language models}.
\newblock \emph{CoRR}, abs/2108.07732.

\bibitem[{Brown et~al.(2020)Brown, Mann, Ryder, Subbiah, Kaplan, Dhariwal,
  Neelakantan, Shyam, Sastry, Askell, Agarwal, Herbert{-}Voss, Krueger,
  Henighan, Child, Ramesh, Ziegler, Wu, Winter, Hesse, Chen, Sigler, Litwin,
  Gray, Chess, Clark, Berner, McCandlish, Radford, Sutskever, and Amodei}]{GPT}
Tom~B. Brown, Benjamin Mann, Nick Ryder, Melanie Subbiah, Jared Kaplan,
  Prafulla Dhariwal, Arvind Neelakantan, Pranav Shyam, Girish Sastry, Amanda
  Askell, Sandhini Agarwal, Ariel Herbert{-}Voss, Gretchen Krueger, Tom
  Henighan, Rewon Child, Aditya Ramesh, Daniel~M. Ziegler, Jeffrey Wu, Clemens
  Winter, Christopher Hesse, Mark Chen, Eric Sigler, Mateusz Litwin, Scott
  Gray, Benjamin Chess, Jack Clark, Christopher Berner, Sam McCandlish, Alec
  Radford, Ilya Sutskever, and Dario Amodei. 2020.
\newblock \href
  {https://proceedings.neurips.cc/paper/2020/hash/1457c0d6bfcb4967418bfb8ac142f64a-Abstract.html}
  {Language models are few-shot learners}.

\bibitem[{Chen et~al.(2021)Chen, Tworek, Jun, Yuan, de~Oliveira~Pinto, Kaplan,
  Edwards, Burda, Joseph, Brockman, Ray, Puri, Krueger, Petrov, Khlaaf, Sastry,
  Mishkin, Chan, Gray, Ryder, Pavlov, Power, Kaiser, Bavarian, Winter, Tillet,
  Such, Cummings, Plappert, Chantzis, Barnes, Herbert{-}Voss, Guss, Nichol,
  Paino, Tezak, Tang, Babuschkin, Balaji, Jain, Saunders, Hesse, Carr, Leike,
  Achiam, Misra, Morikawa, Radford, Knight, Brundage, Murati, Mayer, Welinder,
  McGrew, Amodei, McCandlish, Sutskever, and Zaremba}]{Codex}
Mark Chen, Jerry Tworek, Heewoo Jun, Qiming Yuan, Henrique~Ponde
  de~Oliveira~Pinto, Jared Kaplan, Harrison Edwards, Yuri Burda, Nicholas
  Joseph, Greg Brockman, Alex Ray, Raul Puri, Gretchen Krueger, Michael Petrov,
  Heidy Khlaaf, Girish Sastry, Pamela Mishkin, Brooke Chan, Scott Gray, Nick
  Ryder, Mikhail Pavlov, Alethea Power, Lukasz Kaiser, Mohammad Bavarian,
  Clemens Winter, Philippe Tillet, Felipe~Petroski Such, Dave Cummings,
  Matthias Plappert, Fotios Chantzis, Elizabeth Barnes, Ariel Herbert{-}Voss,
  William~Hebgen Guss, Alex Nichol, Alex Paino, Nikolas Tezak, Jie Tang, Igor
  Babuschkin, Suchir Balaji, Shantanu Jain, William Saunders, Christopher
  Hesse, Andrew~N. Carr, Jan Leike, Joshua Achiam, Vedant Misra, Evan Morikawa,
  Alec Radford, Matthew Knight, Miles Brundage, Mira Murati, Katie Mayer, Peter
  Welinder, Bob McGrew, Dario Amodei, Sam McCandlish, Ilya Sutskever, and
  Wojciech Zaremba. 2021.
\newblock \href {http://arxiv.org/abs/2107.03374} {Evaluating large language
  models trained on code}.
\newblock volume abs/2107.03374.

\bibitem[{Currey et~al.(2017)Currey, Barone, and Heafield}]{TAE}
Anna Currey, Antonio Valerio~Miceli Barone, and Kenneth Heafield. 2017.
\newblock \href {https://doi.org/10.18653/v1/w17-4715} {Copied monolingual data
  improves low-resource neural machine translation}.
\newblock In \emph{Proceedings of the Second Conference on Machine Translation,
  {WMT} 2017, Copenhagen, Denmark, September 7-8, 2017}, pages 148--156.
  Association for Computational Linguistics.

\bibitem[{Devlin et~al.(2019)Devlin, Chang, Lee, and Toutanova}]{BERT}
Jacob Devlin, Ming{-}Wei Chang, Kenton Lee, and Kristina Toutanova. 2019.
\newblock \href {https://doi.org/10.18653/v1/n19-1423} {{BERT:} pre-training of
  deep bidirectional transformers for language understanding}.
\newblock pages 4171--4186.

\bibitem[{Dong and Lapata(2016)}]{LSTMtree}
Li~Dong and Mirella Lapata. 2016.
\newblock \href {https://doi.org/10.18653/v1/p16-1004} {Language to logical
  form with neural attention}.
\newblock In \emph{Proceedings of the 54th Annual Meeting of the Association
  for Computational Linguistics, {ACL} 2016, August 7-12, 2016, Berlin,
  Germany, Volume 1: Long Papers}. The Association for Computer Linguistics.

\bibitem[{Earley(1970)}]{earley:1970}
Jay Earley. 1970.
\newblock An efficient context-free parsing algorithm.
\newblock \emph{Commun. {ACM}}, 13(2):94--102.

\bibitem[{Hendrycks et~al.(2021)Hendrycks, Basart, Kadavath, Mazeika, Arora,
  Guo, Burns, Puranik, He, Song, and Steinhardt}]{APPS}
Dan Hendrycks, Steven Basart, Saurav Kadavath, Mantas Mazeika, Akul Arora,
  Ethan Guo, Collin Burns, Samir Puranik, Horace He, Dawn Song, and Jacob
  Steinhardt. 2021.
\newblock \href {http://arxiv.org/abs/2105.09938} {Measuring coding challenge
  competence with {APPS}}.
\newblock \emph{CoRR}, abs/2105.09938.

\bibitem[{Lewis et~al.(2020)Lewis, Liu, Goyal, Ghazvininejad, Mohamed, Levy,
  Stoyanov, and Zettlemoyer}]{BART}
Mike Lewis, Yinhan Liu, Naman Goyal, Marjan Ghazvininejad, Abdelrahman Mohamed,
  Omer Levy, Veselin Stoyanov, and Luke Zettlemoyer. 2020.
\newblock \href {https://doi.org/10.18653/v1/2020.acl-main.703} {{BART:}
  denoising sequence-to-sequence pre-training for natural language generation,
  translation, and comprehension}.
\newblock In \emph{Proceedings of the 58th Annual Meeting of the Association
  for Computational Linguistics, {ACL} 2020, Online, July 5-10, 2020}, pages
  7871--7880. Association for Computational Linguistics.

\bibitem[{Luong et~al.(2015)Luong, Pham, and Manning}]{Attention}
Thang Luong, Hieu Pham, and Christopher~D. Manning. 2015.
\newblock \href {https://doi.org/10.18653/v1/d15-1166} {Effective approaches to
  attention-based neural machine translation}.
\newblock pages 1412--1421.

\bibitem[{Norouzi et~al.(2021)Norouzi, Tang, and Cao}]{monolingualdata}
Sajad Norouzi, Keyi Tang, and Yanshuai Cao. 2021.
\newblock \href {https://doi.org/10.18653/v1/2021.acl-short.98} {Code
  generation from natural language with less prior knowledge and more
  monolingual data}.
\newblock In \emph{Proceedings of the 59th Annual Meeting of the Association
  for Computational Linguistics and the 11th International Joint Conference on
  Natural Language Processing, {ACL/IJCNLP} 2021, (Volume 2: Short Papers),
  Virtual Event, August 1-6, 2021}, pages 776--785. Association for
  Computational Linguistics.

\bibitem[{Orlanski and Gittens(2021)}]{orlanski-21}
Gabriel Orlanski and Alex Gittens. 2021.
\newblock \href {http://arxiv.org/abs/2106.04447} {Reading stackoverflow
  encourages cheating: Adding question text improves extractive code
  generation}.
\newblock \emph{CoRR}, abs/2106.04447.

\bibitem[{Rabinovich et~al.(2017)Rabinovich, Stern, and Klein}]{ASN}
Maxim Rabinovich, Mitchell Stern, and Dan Klein. 2017.
\newblock \href {https://doi.org/10.18653/v1/P17-1105} {Abstract syntax
  networks for code generation and semantic parsing}.
\newblock In \emph{Proceedings of the 55th Annual Meeting of the Association
  for Computational Linguistics, {ACL} 2017, Vancouver, Canada, July 30 -
  August 4, Volume 1: Long Papers}, pages 1139--1149. Association for
  Computational Linguistics.

\bibitem[{Vaswani et~al.(2017)Vaswani, Shazeer, Parmar, Uszkoreit, Jones,
  Gomez, Kaiser, and Polosukhin}]{Transformer}
Ashish Vaswani, Noam Shazeer, Niki Parmar, Jakob Uszkoreit, Llion Jones,
  Aidan~N. Gomez, Lukasz Kaiser, and Illia Polosukhin. 2017.
\newblock \href
  {https://proceedings.neurips.cc/paper/2017/hash/3f5ee243547dee91fbd053c1c4a845aa-Abstract.html}
  {Attention is all you need}.
\newblock pages 5998--6008.

\bibitem[{Vinyals et~al.(2015)Vinyals, Fortunato, and Jaitly}]{Pointernet}
Oriol Vinyals, Meire Fortunato, and Navdeep Jaitly. 2015.
\newblock \href
  {https://proceedings.neurips.cc/paper/2015/hash/29921001f2f04bd3baee84a12e98098f-Abstract.html}
  {Pointer networks}.
\newblock pages 2692--2700.

\bibitem[{Wang et~al.(1997)Wang, Appel, Korn, and Serra}]{Wang:97}
Daniel~C. Wang, Andrew~W. Appel, Jeffrey~L. Korn, and Christopher~S. Serra.
  1997.
\newblock The zephyr abstract syntax description language.
\newblock In \emph{Proceedings of the Conference on Domain-Specific Languages,
  October 15-17, 1997, Santa Barbara, California, USA}, pages 213--228.

\bibitem[{Yin and Neubig(2017)}]{tranx-1}
Pengcheng Yin and Graham Neubig. 2017.
\newblock \href {https://doi.org/10.18653/v1/P17-1041} {A syntactic neural
  model for general-purpose code generation}.
\newblock In \emph{Proceedings of the 55th Annual Meeting of the Association
  for Computational Linguistics, {ACL} 2017, Vancouver, Canada, July 30 -
  August 4, Volume 1: Long Papers}, pages 440--450. Association for
  Computational Linguistics.

\bibitem[{Yin and Neubig(2018)}]{tranx-2018}
Pengcheng Yin and Graham Neubig. 2018.
\newblock {TRANX}: A transition-based neural abstract syntax parser for
  semantic parsing and code generation.
\newblock In \emph{Proceedings of the 2018 Conference on Empirical Methods in
  Natural Language Processing: System Demonstrations}, pages 7--12, Brussels,
  Belgium. Association for Computational Linguistics.

\end{thebibliography}
\bibliographystyle{acl_natbib}

\appendix

\section{Additional Qualitative Examples}
\label{sec:appendix}
We present examples of code generated by our best models with and without grammar.

\begin{table}[htbp]
\begin{tabular}{|p{1.5cm}|p{5.3cm}| }
 \toprule
Source & \textit{declare an array}\\ \midrule
Gold    & \codeword{my_list = []} \\ \midrule
% Variance natural language length (tokens)&   44  & 98 \\ \hline \hline
Grammar   & \codeword{x = [0] * 2}  \\ \midrule
Without & \codeword{[(0) for _ in range (10000)]}  \\ \midrule
Remark & Source is not precise enough.\\ \bottomrule
\end{tabular}
\end{table}

\begin{table}[htbp]
\begin{tabular}{|p{1.5cm}|p{5.3cm}| }
 \toprule
Source & \textit{increment piece by first element of elt}\\ \midrule
Gold    & \codeword{piece += elt[0]} \\ \midrule
% Variance natural language length (tokens)&   44  & 98 \\ \hline \hline
Grammar   & \codeword{piece += elt[1]}  \\ \midrule
Without & \codeword{piece += elt[1]}  \\ \midrule
Remark & First element of a list is zero. \\ \bottomrule
\end{tabular}
\end{table}

\begin{table}[htbp]
\begin{tabular}{|p{1.5cm}|p{5.3cm}| }
 \toprule
Source & \textit{remove first element of text}\\ \midrule
Gold    & \codeword{text = text[1:]} \\ \midrule
% Variance natural language length (tokens)&   44  & 98 \\ \hline \hline
Grammar   & \codeword{text = text[1:]}  \\ \midrule
Without & \codeword{text[1:}  \\ \midrule
Remark & Syntax mistake for the code without grammar. \\ \bottomrule
\end{tabular}
\end{table}

\begin{table}[htbp]
\begin{tabular}{|p{1.5cm}|p{5.3cm}| }
 \toprule
Source & \textit{get the position of item 1 in `testlist`}\\ \midrule
Gold    & \codeword{[i for i, x in enumerate(testlist) if x == 1]} \\ \midrule
% Variance natural language length (tokens)&   44  & 98 \\ \hline \hline
Grammar   & \codeword{[i for i, v in enumerate(testlist) if v == 1]}  \\ \midrule
Without & \codeword{testlist = [i for i in testlist if i != 1]}  \\ \midrule
Remark & Grammar output is not equal to Gold due to dummy variable. \\ \bottomrule
\end{tabular}
\end{table}

\begin{table}[htbp]
\begin{tabular}{|p{1.5cm}|p{5.3cm}| }
 \toprule
Source & \textit{append a numpy array `b` to a numpy array `a`}\\ \midrule
Gold    & \codeword{np.vstack((a, b))} \\ \midrule
% Variance natural language length (tokens)&   44  & 98 \\ \hline \hline
Grammar   & \codeword{a = numpy.array([b, a])}  \\ \midrule
Without  & \codeword{z = np.array([b]). reshape((3, 3))}  \\ \midrule
Remark & Gold is not accurate with \texttt{np} undefined before. \texttt{vstack} function not in the external vocabulary. \\ \bottomrule
\end{tabular}
\end{table}

\begin{table}[htbp]
\begin{tabular}{|p{1.5cm}|p{5.3cm}| }
 \toprule
Source & \textit{activate is a lambda function which returns None for any argument x.}\\ \midrule
Gold    & \codeword{activate = lambda x : None} \\ \midrule
% Variance natural language length (tokens)&   44  & 98 \\ \hline \hline
Grammar   & \codeword{activate = lambda x = None : x}  \\ \midrule
Without  & \codeword{activate = lambda x : None}  \\ \midrule
Remark & Good BLEU for grammar output while the result is not adequate. \\ \bottomrule
\end{tabular}
\end{table}

\clearpage
\comment{
\begin{table}[htbp]
\begin{tabular}{|p{1.5cm}|p{5.3cm}| }
 \toprule
Source & \textit{convert tuple `t` to list}\\ \midrule
Gold    & \codeword{list(t)} \\ \midrule
% Variance natural language length (tokens)&   44  & 98 \\ \hline \hline
Grammar   & \codeword{[x for x in t for x in t]}  \\ \midrule
Without  & \codeword{[i for i in t]}  \\ \midrule
Remark & Problem of {\sc close} for the Grammar output. Without grammar the code is correct but with a low BLEU.\\ \bottomrule
\end{tabular}
\end{table}}
\vfill

\end{document}